%% file: emnlp2022_review.tex
\pdfoutput=1

\documentclass[11pt]{article}

\usepackage[]{EMNLP2022}

\usepackage{times}
\usepackage{latexsym}

\usepackage[T1]{fontenc}

\usepackage[utf8]{inputenc}

\usepackage{microtype}

\usepackage{inconsolata}

\title{LUNA: Language Understanding with Number Augmentations on Transformers via Number Plugins and Pre-training}
\makeatletter
\newcommand{\printfnsymbol}[1]{%
  \textsuperscript{\@fnsymbol{#1}}%
}
\makeatother

\author{
Hongwei Han\textsuperscript{\rm 1}\thanks{\indent The contributions by Hongwei Han, Jialiang Xu and Yijia Shao have been conducted and completed during their internships at Microsoft Research Asia, Beijing, China.}~\thanks{\indent Equal Contribution.} ,
Jialiang Xu\textsuperscript{\rm 2}\printfnsymbol{1}\printfnsymbol{2},
Mengyu Zhou\textsuperscript{\rm 3}\thanks{\indent Corresponding author.}, 
Yijia Shao\textsuperscript{\rm 4}\printfnsymbol{1}, 
Shi Han\textsuperscript{\rm 3}, 
Dongmei Zhang\textsuperscript{\rm 3} \\
\textsuperscript{\rm 1} Tsinghua University,
\textsuperscript{\rm 2} University of Illinois at Urbana-Champaign \\
\textsuperscript{\rm 3} Microsoft Research,
\textsuperscript{\rm 4} Peking University \\
\texttt{\href{mailto:hhw20@mails.tsinghua.edu.cn}{hhw20@mails.tsinghua.edu.cn}}, \\
\texttt{\href{mailto:jx17@illinois.edu}{jx17@illinois.edu}}, \texttt{\href{mailto:shaoyj@pku.edu.cn}{shaoyj@pku.edu.cn}},\\ \texttt{\{\href{mailto:mezho@microsoft.com}{mezho}, \href{mailto:shihan@microsoft.com}{shihan}, \href{mailto:dongmeiz@microsoft.com}{dongmeiz}\}@microsoft.com}}

\input{tools}

\begin{document}
\maketitle
\begin{abstract}
\input{0-abstract}
\end{abstract}

\section{Introduction}
\input{1-intro}

\section{Related Work}
\input{2-related}

\section{Methods}
\input{3-method}

\section{Experiments}
\input{4-experiments}



\section{Conclusion}
\input{7-conclusion}

\section*{Limitations}
\input{5-discussion}

\section*{Ethics Statement}
\input{ethic}

\bibliography{anthology,custom}
\bibliographystyle{acl_natbib}

\appendix
\input{appendix}

\end{document}

%% file: tools.tex
\usepackage{subfig}
\usepackage{booktabs}
\usepackage{nicefrac}       
\usepackage{xspace}
\usepackage{cprotect}
\usepackage{xcolor}
\usepackage{appendix}
\usepackage{enumitem}
\usepackage{balance}
\usepackage{circledsteps}
\usepackage{algorithm}
\usepackage{algpseudocode}
\usepackage{graphicx}
\usepackage{amsmath}
\usepackage{amssymb}
\usepackage{multirow}

\usepackage{ulem}

\newcommand{\reffig}[1]{Figure~\ref{#1}}
\newcommand{\refsec}[1]{\S\ref{#1}} 
\newcommand{\reftab}[1]{Table~\ref{#1}}

\newcommand{\grey}[1]{\textcolor{gray}{#1}}

\def\eg{\textit{e.g.}\xspace}
\def\Eg{\textit{E.g.}\xspace}

\def\etc{\textit{etc.}\xspace}
\def\ie{\textit{i.e.}\xspace}
\def\vs{\textit{vs.}\xspace}

\newcommand{\cmark}{$\checkmark$}%
\newcommand{\xmark}{$\times$}%
\newcommand{\rx}[2]{$#1_{#2}$}
\newcommand{\rxo}[1]{$#1$}
\newcommand{\ri}[2]{$\underline{\textbf{#1}}_{#2}$}
\newcommand{\rio}[1]{$\underline{\textbf{#1}}$}
\newcommand{\rii}[2]{$\underline{#1}_{#2}$}
\newcommand{\riio}[1]{$\underline{#1}$}


%% file: 0-abstract.tex


Transformers are widely used in NLP tasks. However, current approaches to leveraging transformers to understand language expose one weak spot: Number understanding. In some scenarios, numbers frequently occur, especially in semi-structured data like tables. But current approaches to rich-number tasks with transformer-based language models abandon or lose some of the numeracy information -- \eg, breaking numbers into sub-word tokens -- which leads to many number-related errors. In this paper, we propose the LUNA framework which improves the numerical reasoning and calculation capabilities of transformer-based language models. With the number plugin of NumTok and NumBed, LUNA represents each number as a whole to model input. With number pre-training, including regression loss and model distillation, LUNA bridges the gap between number and vocabulary embeddings. To the best of our knowledge, this is the first work that explicitly injects numeracy capability into language models using Number Plugins. Besides evaluating toy models on toy tasks, we evaluate LUNA on three large-scale transformer models (RoBERTa, BERT, TabBERT) over three different downstream tasks (TAT-QA, TabFact, CrediTrans), and observe the performances of language models are constantly improved by LUNA. The augmented models also improve the official baseline of TAT-QA ($EM: 50.15 \rightarrow 59.58$) and achieve SOTA performance on CrediTrans ($F1=86.17$).

%% file: 1-intro.tex
\begin{figure*}[t]
  \centering
  \vspace{-5mm}
  \includegraphics[width=\textwidth]{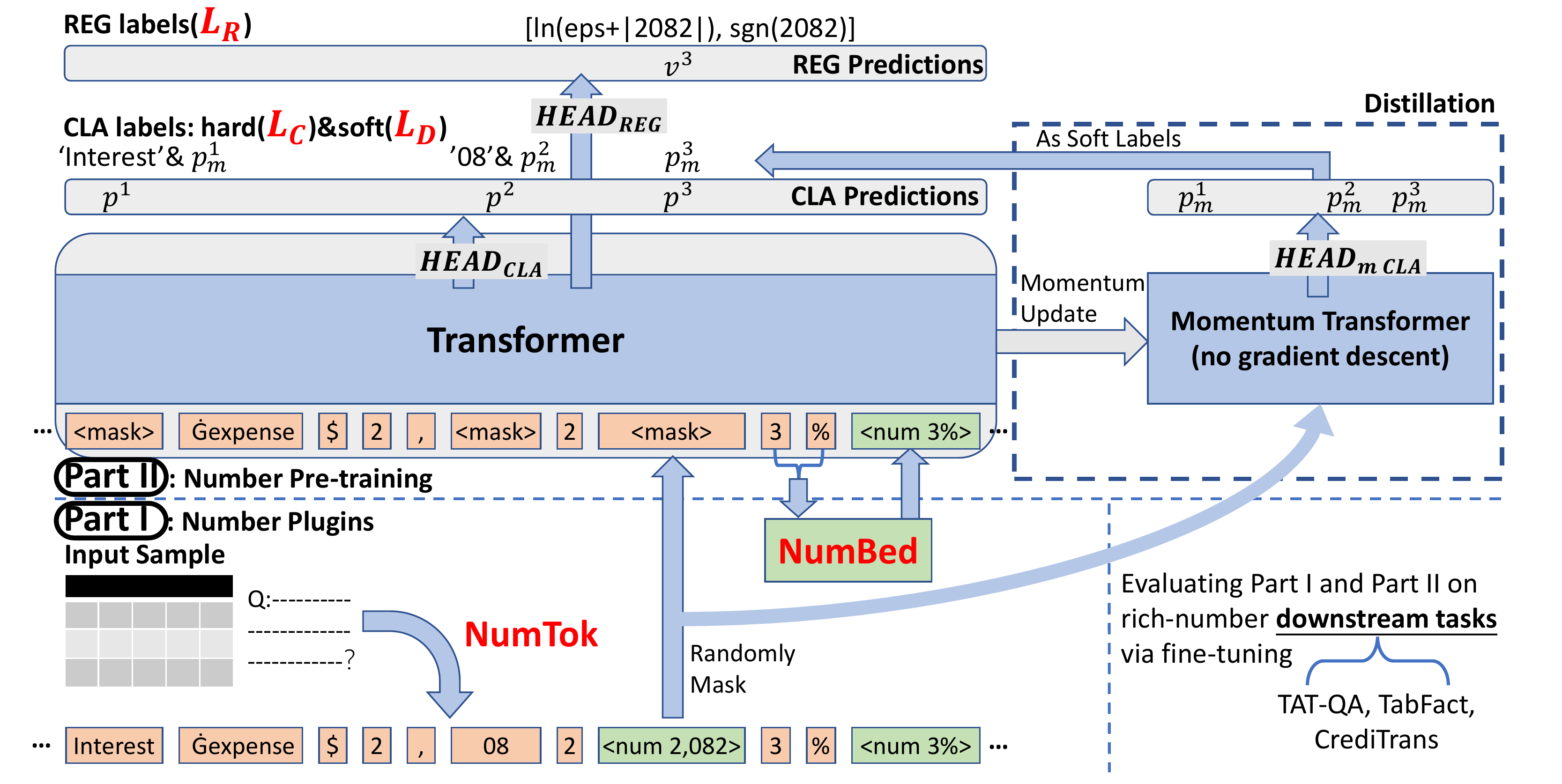}
  \caption{Overview of LUNA Framework. \Circled{Part I} number plugins (introduced in \refsec{sec:numtok}) include NumTok (number tokenizer) and NumBed (number embedder). \Circled{Part II} number pre-training (introduced in \refsec{sec:num-pretrain}). $\mathcal{L}_D$, $\mathcal{L}_R$, and $\mathcal{L}_C$ respectively denote distillation loss (on soft labels), regression loss (on real values), and the original classification loss (classifying the masked tokens, on hard labels). $HEAD_{REG}$ and $HEAD_{CLA}$ are learnable modules that respectively project the last hidden states into 2 and vocab-size dimensions. The transformer and momentum transformer are initialized with the same checkpoint. }
  \label{fig:luna-overview}
  \vspace{-5mm}
\end{figure*}

\label{sec:intro}

\begin{table}[t]
    \centering
    \includegraphics[width=\linewidth]{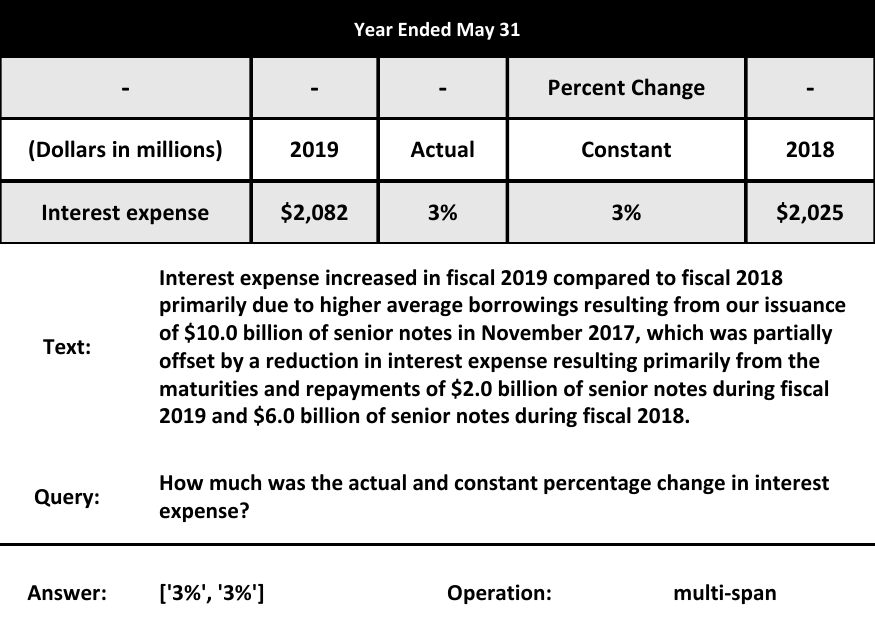}
    \caption{TAT-QA Example: A Financial Report Table.}
    \label{tab:A}
    \vspace{-5mm}
\end{table}

Numbers are common in everyday NLP scenarios. Transformer-based language models~\citep{plm-survey} are popular for number-rich tasks such as TAT-QA~\citep{zhu-etal-2021-tat}, TabFact~\citep{chen2020TabFact} and CrediTrans~\citep{padhi2021tabular} (see \refsec{sec:datasets} for details). For example, \reftab{tab:A} comes from a financial report in TAT-QA. To correctly understand the question and give the answer, a model must understand the semantics of each number in this example. Current approaches to TAT-QA (see \refsec{sec:datasets}) often serialize the (table, text, query) triplet into a sequence, and then feed it into a transformer-based language model such as RoBERTa.


The weak spot of number understanding of transformers is magnified when facing rich-number tasks (\eg, the above QA, fact check, time series modeling tasks) that heavily depends on numerical reasoning and calculation. Errors could easily occur by adopting existing number processing and representation approaches in NLP (\eg, a bad case for \reftab{tab:A} in \refsec{sec:empirical-case}). There are two great challenges: tokenization and representation of numbers. 

Recent studies have already shown that current \textbf{tokenization} methods for numbers in language are suboptimal and sporadic~\citep{thawani-etal-2021-representing}: Numbers are either filtered out, collapsed into \texttt{<unk>} token, quantized into finite bins, or split into arbitrary subword tokens. In other words, numeracy information is abandoned or scattered around. \Eg, ``3.1415'' is split into ``3'', ``.'', ``14'', ``15'' by BPE~\citep{sennrich-etal-2016-neural} (used in RoBERTa) and into ``3'', ``.'', ``141'', ``\#\#5'' by WordPiece~\citep{wu2016wordpiece}.


On the other hand, to augment the numerical reasoning and calculation capability of language models, previous work such as Time2Vec~\cite{time2vec} and DICE~\cite{sundararaman-etal-2020-methods} design value-based feature vectors for number \textbf{representation}. However, those methods are not evaluated on transformers or real-world tasks (as discussed in \refsec{sec:exp}, those methods indeed hurt). 


To better solve the number issues, we propose \textbf{LUNA} (\textbf{L}anguage \textbf{U}nderstanding with \textbf{N}umber \textbf{A}ugmentations on Transformers via Number Plugins and Pre-training) framework. As shown in \reffig{fig:luna-overview}, LUNA works as a patch for existing language models when handling rich-number tasks. Through tough and massive attempts, we also exclude many failed designs like number prompts (as discussed in \textbf{Limitations} section), obtaining the simple and effective LUNA. 

Our first key idea behind LUNA is that: \textit{Each number should be represented as a whole for model input} (rather than broken into subword tokens or quantized into binned tokens). Learning embedding representation for raw number strings properly could help exploit numeracy information for downstream tasks.
In LUNA, this idea corresponds to \Circled{Part I} number plugin in \reffig{fig:luna-overview}. Each number in the input (table and text) string will be first recognized by our \textbf{NumTok} (number tokenizer, see \refsec{sec:numtok}) and inserted as a special \texttt{<num ?>} (? denotes the recognized number string) token into the original input token sequence. Then, the embedding of \texttt{<num ?>} is computed by our \textbf{NumBed} (number embedder, see \refsec{sec:numbed}) which encodes the recognized number string as an embedding vector of the same dimension as the token embedding of the transformer. In detail, for Numtok, we discuss two designs: AddBack and Replace, which keep or delete the original representation of numbers, as will be introduced in \refsec{sec:numtok}. And for Numbed, we compare CharLSTM and CharFormer which encode the char sequence of the number string with an LSTM or a transformer, and in \refsec{sec:numbed}, we further compare different NumBed model sizes.

The newly introduced number embedding may be incompatible with the embedding space of the model's vocabulary tokens. To bridge the gap between number and vocabulary embeddings, \Circled{Part II} of LUNA -- \textbf{number pre-training} on data from downstream tasks (see \refsec{sec:num-pretrain}) -- is designed for either starting the language model from scratch or from a pre-trained checkpoint. First, to incorporate the \texttt{<num ?>} token, we modified the classical MLM (masked language model) objective with specially designed regression loss for it. Then, when starting from a pre-trained checkpoint, it is easy for the model to forget existing knowledge and overfit. To prevent this, a teacher-student model distillation process is designed as a regularization constraint, and this paper is the first work to leverage model distillation in number pre-training. 

By applying LUNA to existing transformer-based language models, we observe constant performance improvements on QA, fact-check, and time-series tasks (see \refsec{sec:results}) over three transformer-based language models. We evaluate RoBERTa~\citep{liu2019roberta} on TAT-QA~\citep{zhu-etal-2021-tat}, BERT~\citep{bert} on TabFact~\citep{chen2020TabFact}, and TabBERT~\citep{padhi2021tabular} on CrediTrans. With the help of LUNA framework, RoBERTa beats the baseline ($EM: 50.15 \rightarrow 59.58$, and $EM_{num}: 38.37 \rightarrow 63.33$) on TAT-QA. TabBERT reaches new SOTA numbers on CrediTrans ($F1=86.17$, the previous SOTA is 84.79). By further 
analyzing the dataset and conducting empirical studies, we further demonstrate how LUNA helps preserve numeracy information for downstream tasks. According to the controlled experiments, we find that: 1) For NumTok and NumBed choices, AddBack is much better than Replace, CharLSTM is an overall good choice, and a larger numbed model size is better. 2) For number pre-training, distillation loss usually helps and regression loss only helps on tagging tasks. These conclusions benefit future model design and selection. 

In summary, our major contributions are:
\begin{itemize}
    \item To the best of our knowledge, our work first attempts to explicitly deal with the number issues on existing transformer-based language models. Besides evaluating on toy tasks, we evaluate the numeracy ability of our designs on real tasks as well. 
    \item We propose LUNA framework to enhance the number understanding capabilities of language models. The code and data of LUNA are open-sourced at \url{https://github.com/zmy/LUNA}
    \item We do controlled experiments to discuss NumTok\&NumBed choices and number pre-training objectives. 
\end{itemize}

%% file: 2-related.tex
\label{sec:related} 

\subsection{Rich-number Tasks}
\label{sec:datasets}

In this paper, we take three typical rich-number datasets and corresponding tasks as examples.

TAT-QA\footnote{Tabular And Textual dataset for Question Answering, \url{https://nextplusplus.github.io/TAT-QA/}} \citep{zhu-etal-2021-tat} is a recent QA dataset requiring numerical reasoning over realistic tabular and textual data. In total, TAT-QA contains 16,552 questions associated with 2,757 hybrid (table + paragraphs) contexts from real-world financial reports. 20\% of the hybrid contexts are used as development and test sets. A table in TAT-QA has 3$\sim$30 rows and 3$\sim$6 columns with at least 2 human-verified relevant nearby paragraphs.

TabFact\footnote{TabFact, \url{https://tabfact.github.io/}} \citep{chen2020TabFact} is a fact verification dataset with 16k Wikipedia tables as evidence for 118k human-annotated statements. In TabFact, for each Wikipedia table (with caption), the goal is to distinguish which given statements are entailed by the table and which are refuted by it. A table in TabFact has less than 50 rows and 10 columns. 

CrediTrans\footnote{Credit Card Transaction Dataset, \url{https://ibm.ent.box.com/v/tabformer-data}}\citep{padhi2021tabular} is a time series corpus for credit card transactions with 24 million transactions from 20,000 users. Each transaction (row) has 12 fields (columns) consisting of both continuous and discrete attributes, such as merchant name, merchant address, transaction amount, etc. We can see an example of TAT-QA over \reftab{tab:A}, the one of TabFact and CrediTrans over \reftab{tab:B} and \reftab{tab:C} in appendix. The numeracy statistics are collected in \refsec{sec:number-stat} in appendix as well.

Math Word Problems~\citep{mwp} are another set of rich-number tasks besides tabular tasks. We leave it as future work because neural-based math-word-problem solvers are often generation models, which need the capability of decoding numbers. We focus this paper on understanding instead of generation. 


\subsection{Numeracy in NLP}
In recent years, some effort has been made in boosting numeracy capability in NLP. Position Embedding~\citep{transformer} embeds integers into vectors, and as an extension, Time2Vec~\citep{time2vec} and DICE~\citep{sundararaman-etal-2020-methods} embed float numbers into vectors. TabBERT~\citep{padhi2021tabular} quantifies numbers into buckets by their value. This might be a good attempt, but loses the information of similarity between nearby numbers in different buckets. These approaches embed numbers based on their values. DICE is famous for beating many methods on toy tasks like predicting +/- of two numbers, and we treat it as a baseline of NumBed choices. We find that DICE is not a good choice for real tasks\footnote{We don't try Time2Vec, because numbers are not periodic. }, and even not good on toy tasks when measured with different metrics, as discussed in \refsec{sec:exp}. 

\label{sec:tab-models}

In most cases, numbers are tokenized in language ways in transformer-based language models. The common practice of language models usually adopts a fixed vocabulary and the sub-word tokenization approaches like BPE tokenizer~\citep{sennrich-etal-2016-neural} in RoBERTa and WordPiece tokenizer~\citep{wu2016wordpiece} in BERT. Instead of considering the number as an ensemble, the tokenizer splits the number into arbitrary tokens. NumBERT~\citep{numbert} is pre-trained from scratch over a modified dataset where all numbers have been replaced with scientific notation (e.g., replacing 314.1 with 3141[EXP]2), in order to help the model understand the significance and exponent of numbers. However, NumBERT still breaks a number into sub-word pieces. As we have discussed in \refsec{sec:intro}, breaking a number into sub-word pieces or quantifying it is not a grounded practice. A bad case analysis is provided in \refsec{sec:empirical-case}. 

Some previous work also focuses on number decoding~\citep{spithourakis-riedel-2018-numeracy}.

%% file: 3-method.tex
\label{sec:method} 

As shown in \reffig{fig:luna-overview}, there are two major parts in LUNA:
Part I, number plugin modules -- NumTok (see \refsec{sec:numtok}) and NumBed (see \refsec{sec:numbed}) -- which change the input to a language model such as RoBERTa or BERT.
Part II, number pre-training (part II, see \refsec{sec:num-pretrain}) from scratch or from an existing checkpoint of the language model. 

In LUNA, we take the following steps to enhance a language model:
First, NumTok changes the result token sequence of the model's tokenizer by inserting a special \texttt{<num ?>} (? denotes the recognized number string) token for each number.
Second, the embedding of \texttt{<num ?>} is computed by NumBed which encodes the recognized number string as an embedding vector of the same dimension as the word-embedding. Third, number pre-training is taken place to align the space of number embedding and that of vocabulary embedding. This will also allow the model to fit unlabeled downstream data.
Finally, the common fine-tuning process on the language model (with its additional task-specific modules) is taken for a downstream task.

\subsection{Number Plugins}
\label{sec:numtok}

\begin{figure}
    \centering
    \includegraphics[width=\linewidth]{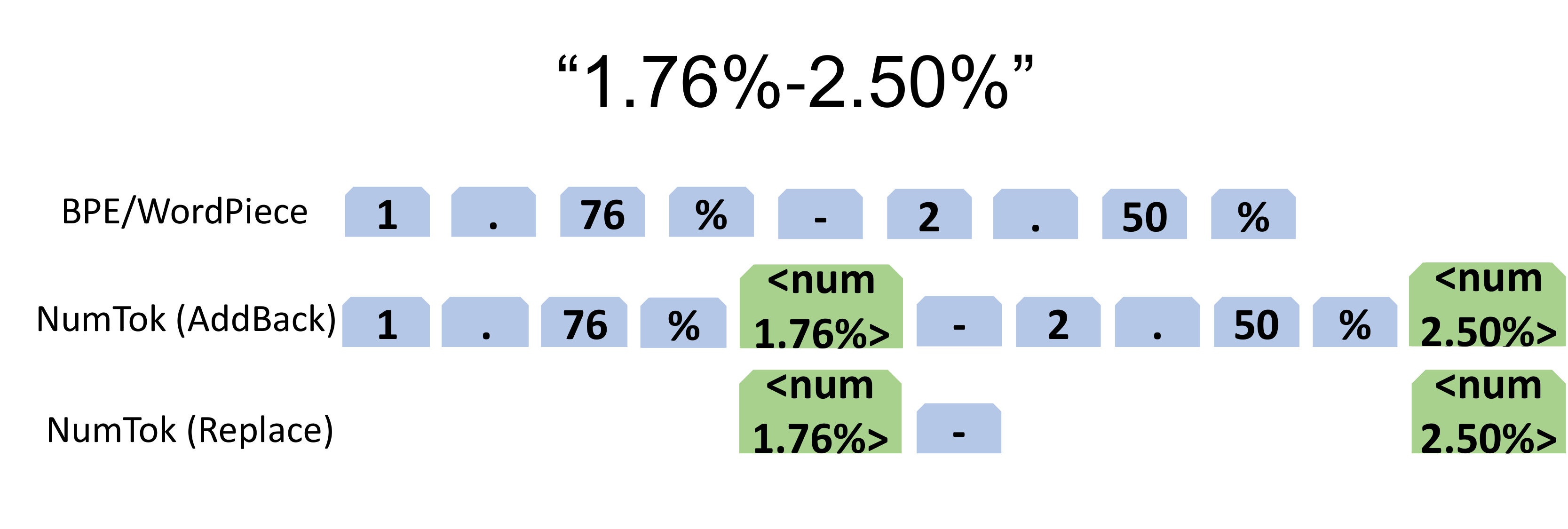}
    \caption{The Two Options of NumTok, Compared with BPE/WordPiece. }
    \label{fig:numtok-demo}
    \vspace{-5mm}
\end{figure}
The first step (part I) in LUNA is Number Plugins. 

\textbf{NumTok} (as a wrapper) slightly changes the behavior of a model's default tokenizer by inserting ``<num>'' into an input string at locations where a number appears. Then, ``<num>'' will be recognized as a new special token \texttt{<num ?>} by NumTok in its output. As shown in \reffig{fig:numtok-demo}, there are two ways \texttt{<num ?>} to be inserted by NumTok: \textbf{AddBack} and \textbf{Replace}. The former adds \texttt{<num ?>} to the end of the original vocabulary token(s) of the number. The latter fully replaces these token(s).

The key algorithm in NumTok is recognizing numbers in the input string. We adopt a simple but effective filter-split-check approach to ensure characters in the same number are recognized as a whole, while irrelevant characters are excluded. \Eg, ``1.76\%-2.50\%'' should be recognized as two positive percentages ``1.76\%'' and ``2.50\%''. Regular expressions used by NumTok are listed in \refsec{app:numtok_re}. 



\label{sec:numbed}

Then, we should assign each \texttt{<num ?>} token an embedding vector that could be used for operations such as attention with other vocabulary tokens. For this, we propose \textbf{NumBed}, which is a trainable neural network aiming at generating embedding reflecting numeracy properties of the original number. As shown in \reffig{fig:luna-overview}, the number (\texttt{<num ?>} token) embedding is the generated NumBed embedding, which will be part of the input sequence to the language model. 


For the model design of NumBed, there is a wide range of choices, some provided by previous work like DICE~\citep{sundararaman-etal-2020-methods} and some designed by us: 
\textbf{CharLSTM} embeds the number on character level, \ie the characters in the number string are first embedded with one-hot lookup embedding and then passed into a Bidirectional LSTM model. The underlying intuition is that 1) semantics in numbers could be queried either left-to-right or right-to-left, and 2) LSTM naturally encodes positional order of the digits, which mimics the definition of number value (the sum of numbers on all digits, multiplied by its digit significance). The characteristic embedding is obtained by averaging the final hidden state of all LSTM layers.
\textbf{CharFormer} utilizes a transformer encoder layer and rule-
based positional encoding. Characters in the number string are
first embedded with one-hot lookup embedding and then added by
a positional embedding indicating its digit significance (i.e. the
relative position w.r.t. the decimal point). The result embedding
sequence is then sequentially passed through the transformer encoder layer and an LSTM model. The characteristic embedding is
obtained by averaging the final hidden state of all LSTM layers.

\subsection{Number Pre-training}
\label{sec:num-pretrain}

The next step (part II) in LUNA is number pre-training on downstream data. There are two reasons to do so: First, randomly initialized \texttt{<num ?>} embedding could damage the distribution of original input embeddings. Pre-training could help bridge the gap between number and vocabulary embeddings. Second, by pre-training on data of downstream tasks, the model can learn better the domain distributions of the tasks compared to directly fine-tuning~\cite{gururangan-etal-2020-dont}.

\subsubsection{Pre-training from Checkpoints}
\label{sec:checkpoint}
Reproducing the original pre-training process of a language model can be very costly. Thus, the first option in LUNA part II is to conduct intermediate pre-training from a pre-trained checkpoint (like RoBERTa and BERT) of the model.

The classical mask language model (MLM) objective~\cite{bert} is adapted for LUNA. 
We keep the common practice that overall 
15\% of the input tokens are masked for recovery at the corresponding positions of the model output. For 80\% of them, the input tokens are replaced by a \texttt{[MASK]} token. For 50\% of the rest, they are replaced by a randomly selected token from the vocabulary; For the other 50\%, they keep what they are.

For a masked \texttt{<num ?>} token, a newly designed regression supervision is applied to fit the log absolute and the sign of its corresponding numeric value. And for a masked text token, classification supervision is used. To be exact, let $k$ denote the number of masked tokens in a mini-batch (about 15\% of total), $\mathcal{N}=\{(o,v)\}$ denote the set of \textbf{$o$}utput vector and real \textbf{$v$}alue of masked \texttt{<num ?>} tokens, $\mathcal{T}=\{(o,t)\}$ denote the set of \textbf{$o$}utput vector and original \textbf{$t$}oken of masked vocabulary tokens. Obviously, $|\mathcal{N}|+|\mathcal{T}|=k$. Let a learnable MLP $HEAD_{REG}$ denote the regression head and $HEAD_{CLA}$ denote the classification head. Then we can define MLM loss ($\mathcal{L}_{MLM}$) as: 
\begin{equation*}
\begin{split}
    &\mathcal{L}_{MLM}=\frac{1}{k}(\mathcal{L}_{REG}+\mathcal{L}_{CLA})\\
    &\mathcal{L}_{REG}=\sum_{\mathcal{N}}MSE(HEAD_{REG}(o),\\
    &\hspace{32mm}[ln(eps+|v|),sgn(v)])\\
    &\mathcal{L}_{CLA}=\sum_{\mathcal{T}}CE(HEAD_{CLA}(o),onehot(t))
\end{split}
\end{equation*}
Here the mean-squared-error regression loss ($\mathcal{L}_{REG}$ or $\mathcal{L}_{R}$) takes in $ln(eps+|v|)$ to reflect actual range rather than exact value. The classification loss ($\mathcal{L}_{CLA}$ or $\mathcal{L}_{C}$) is the usual cross entropy.

Since downstream datasets are usually far smaller than the original pre-train datasets of the models, the overfitting problem may happen during the LUNA number pre-training phase. To solve the problem, we adopt a regularization method called model distillation~\cite{li2021ALBEF}.

During model distillation, there is a student network $Net$ and a teacher network (also called momentum network) $Net_m$. They both start with the same parameter initialization (from a checkpoint), take the same masked embedding sequence as input, and generate probability distribution vectors ($p$ by $Net$ and $p_m$ by $Net_m$) over the vocabulary. Please note that there is not a so-called distillation head, and $p$ is projected by $HEAD_{CLA}$. 
When treating $p_m$ as soft label (different from the hard label $t$ in $\mathcal{L}_{C}$), the distillation loss ($\mathcal{L}_{distill}$ or $\mathcal{L}_{D}$) is defined as: $$\mathcal{L}_{distill}=\frac{1}{k}\sum_{\mathcal{N}+\mathcal{T}}CE(p,p_m)$$
When applying $\mathcal{L}_{D}$, $Net$ is updated via gradient, but $Net_m$ is updated only via a fraction of $Net$. In other words, $Net_m$ is the momentum accumulation of $Net$. Let $\tau$ denote the momentum coefficient (default set to 0.995). At each step, 
$Net_m \gets \tau Net_m+(1-\tau) Net$.

The total pre-train loss ($\mathcal{L}_{pre-train}$) is defined as (where $\alpha$ is a warm-up coefficient which grows with the training steps): $$\mathcal{L}_{pre-train}=(1-\alpha)\mathcal{L}_{MLM}+\alpha\mathcal{L}_{distill}$$

Overall, $HEAD_{CLA}$ is under the supervision of $\mathcal{L}_{C}$ and $\mathcal{L}_{D}$, $HEAD_{REG}$ is under the supervision of $\mathcal{L}_{R}$, and $HEAD_{m~CLA}$ is the momentum accumulation of $HEAD_{m~CLA}$. 

\subsubsection{Pre-training from Scratch}
\label{sec:scratch}
When there are enough resources, one can still re-run the original pre-training process of a language model (like TabBERT) with LUNA number plugins and MLM loss ($\mathcal{L}_{MLM}$, which includes the regression loss) without model distillation. However, pre-training from scratch is not the focus of this paper and we just want to verify the effectiveness of LUNA in different situations. 




%% file: 4-experiments.tex
\label{sec:exp} 
\subsection{Datasets and Evaluation Metrics}
\label{sec:exp-data}
In \refsec{sec:related}, we have already discussed several language models (RoBERTa, BERT, TabBERT) and their corresponding tasks (TAT-QA, TabFact, CrediTrans). Both part I and part II of the LUNA framework will be evaluated on these models and tasks. For simplicity, in this paper, we only implemented limited model\&task combinations: RoBERTa on TAT-QA, BERT on TabFact, and TabBERT on CrediTrans.

\subsubsection{Number Pre-training Datasets}
\label{sec:pretrain-dataset}
As discussed in \refsec{sec:num-pretrain}, before fine-tuning on the aforementioned tasks, in LUNA a language model will be pre-trained with rich-number data.


When pre-training from the checkpoint with tuned parameters, MLM loss will be added together with distillation loss (see \refsec{sec:checkpoint}). For RoBERTa and BERT running on text and table inputs, we collect unlabeled table-text pairs (which are serialized into sequences) from TAT-QA and WikiTables (where TabFact comes from). When pre-training from scratch with randomly initialized parameters, MLM with regression loss is adopted in LUNA (see \refsec{sec:scratch}). In our experiments, we pre-train TabBERT (with number plugin) from scratch on CrediTrans samples. Please see \refsec{app:number-datasets} for more details about pre-training datasets. 

\begin{table}[t]
\centering
\resizebox{\linewidth}{!}{%
\begin{tabular}{cccccccc} 
\toprule
           & \multicolumn{2}{c}{Decoding} & \multicolumn{2}{c}{Addition} & \multicolumn{2}{c}{Subtraction} & List Max  \\ 
\cmidrule(lr){2-3}\cmidrule(lr){4-5}\cmidrule(lr){6-7}\cmidrule(lr){8-8}
           & Sig$\downarrow$ & Exp                    & Sig$\downarrow$ & Exp                     & Sig$\downarrow$ & Exp                       & Acc       \\ 
\cmidrule(lr){2-8}
CharLSTM & \rio{0.0946}     & \rio{99.97\%}                        & \rio{0.5572}     &  \rio{99.46\%}           & \rio{1.367}     & \rio{97.17\%} &  \rio{98.55\%}         \\
CharFormer &\riio{0.1799}&\riio{99.85\%}&\riio{0.6718}&\riio{99.44\%}&\riio{1.546}&\riio{96.86\%}&\riio{97.71\%}         \\
RoBERTa    & \rxo{2.400}    & \rxo{34.75\%}     & \rxo{2.332}     & \rxo{37.80\%}                         & \rxo{4.166}     & \rxo{52.16\%} & \rxo{32.84\%}           \\
DICE       & \rxo{2.309}    &  \rxo{67.90\%} & \rxo{2.453}     & \rxo{69.29\%}                         & \rxo{3.819}    &  \rxo{39.53\%}  & \rxo{92.45\%}           \\
\bottomrule
\end{tabular}
}
\caption{Probing numeracy on a number dataset constructed from numbers in real-world tables. We report the RMSE value for all significand (sig) predicting tasks, and classification accuracy for all exponent (exp) predicting tasks and the list maximum task. $\downarrow$ denotes lower is better. We use the same probing models as in \citet{sundararaman-etal-2020-methods}. }
\label{tab:numeracy}
\vspace{-5mm}
\end{table}

\begin{table*}
\centering
\vspace{-5mm}
\resizebox{0.85\textwidth}{!}{%
\begin{tabular}{lcccccc|ccccc|c} 
\toprule
    & \multicolumn{6}{c|}{LUNA Choices}                                                             & \multicolumn{3}{c}{RoBERTa} & \multicolumn{2}{c|}{BERT}&  \\
\cmidrule(lr){2-7}\cmidrule(lr){8-10}\cmidrule(lr){11-12}
    & \multicolumn{3}{c}{Part I}                    & \multicolumn{3}{c|}{Part II}                  & \multicolumn{3}{c}{TAT-QA} & \multicolumn{2}{c|}{TabFact} & \\ 
\cmidrule(lr){2-4}\cmidrule(lr){5-7}\cmidrule(lr){8-10}\cmidrule(lr){11-12}
    & NumTok          & NumBed          & +Size     & $\mathcal{L}_{D}$ & $\mathcal{L}_{R}$     & $\mathcal{L}_{C}$     & $EM$ & $F1$ & $EM_{num}$  & $ACC$ & $ACC_{cx}$ & AVG   \\ 
\midrule
\Circled{0} & AddBack        & CharLSTM        & 9M        & \cmark        & \cmark        & \cmark        & \ri{59.58}{2.7}    & \ri{67.15}{2.7}     & \ri{63.33}{6.4}             &  \rx{66.07}{0.5}       & \rx{62.24}{0.4} &  \rio{63.67}                 \\
\Circled{1} & Replace       & \grey{CharLSTM} & \grey{9M} & \grey{\cmark} & \grey{\cmark} & \grey{\cmark} & \rx{51.87}{1.7}      & \rx{59.94}{1.6}      & \rx{50.21}{4.0}      & \rx{65.41}{0.5}       & \rx{61.43}{0.6}    & \rxo{57,77}\\
\Circled{2} & \grey{AddBack} & CharFormer        & \grey{9M} & \grey{\cmark} & \grey{\cmark} & \grey{\cmark} & \rx{57.03}{4.0}      & \rx{64.83}{3.8}      & \rx{57.48}{9.8}              & \rii{66.81}{0.1}       &     \rii{62.58}{0.2}    &  \rxo{61.75}               \\
\Circled{3} & \grey{AddBack}  & DICE   & \grey{\xmark}    & \grey{\cmark}        & \grey{\cmark}        & \grey{\cmark}        & 
\rx{47.15}{1.2}      & \rx{54.93}{1.0}      & \rx{35.39}{3.1}              & \rx{62.61}{1.2}       & \rx{59.51}{0.6}  &\rxo{51.92}                 \\
\Circled{4} & \grey{AddBack} & \grey{CharLSTM} & 1M        & \grey{\cmark} & \grey{\cmark} & \grey{\cmark} & \rii{57.86}{2.7}      & \rii{65.56}{2.6}      & \rii{58.41}{6.4}              & \rx{65.73}{0.1}       &            \rx{61.79}{0.3}         & \riio{61.87}           \\
\Circled{5} & \grey{AddBack} & \grey{CharLSTM} & 0.1M      & \grey{\cmark} & \grey{\cmark} & \grey{\cmark} & \rx{56.99}{2.0}      & \rx{64.96}{1.7}      & \rx{57.81}{4.5}              &\rx{65.40}{0.7}       &     \rx{61.67}{0.6}  & \rxo{61.37}            \\
\Circled{6} & \grey{AddBack} & \grey{CharLSTM} & \grey{9M} & \grey{\cmark} & \xmark        & \grey{\cmark} & \rx{55.51}{2.9}      & \rx{63.62}{2.8}      & \rx{53.75}{5.3}              & \ri{67.23}{0.7}     & \ri{62.94}{0.5}                    & \rxo{60.61}                \\
\Circled{7} & \grey{AddBack} & \grey{CharLSTM} & \grey{9M} & \xmark        & \grey{\cmark} & \grey{\cmark} & \rx{54.33}{2.3}      & \rx{62.13}{2.0}      & \rx{51.73}{5.5}              &  \rx{65.40}{0.3}      &  \rx{62.19}{0.1}                   & \rxo{59.16}             \\
\Circled{8} & \grey{AddBack} & \grey{CharLSTM} & \grey{9M} & \xmark        & \xmark        & \xmark        & \rx{52.05}{2.8}      & \rx{59.92}{2.8}      & \rx{44.75}{6.7}              &  \rx{62.41}{1.8}       & \rx{59.87}{1.1}                    & \rxo{55.80}                 \\
\Circled{X} & \xmark        & \xmark          & \xmark    & \xmark        & \xmark        & \xmark        & \rx{50.15}{0.8}      & \rx{57.84}{0.7}      & \rx{38.37}{1.9}              & \rx{62.13}{0.6}       & \rx{59.71}{0.4}                    &\rxo{53.64}                 \\
\bottomrule
\end{tabular}
}
\caption{LUNA Evaluations on Language Models (RoBERTa, BERT) and Downstream Tasks (TAT-QA, TabFact). The last column AVG is the average of the five metrics, we add this column to demonstrate the overall capability of each controlled experiment. }
\label{tab:luna-eval}
\vspace{-5mm}
\end{table*}


\begin{table}[t]
\centering
\resizebox{0.45\textwidth}{!}{%
\begin{tabular}{cccccc|c} 
\toprule
    &\multicolumn{5}{c|}{LUNA Choices}  & \multicolumn{1}{c}{CrediTrans} \\
\cmidrule(lr){2-4}\cmidrule(lr){5-6}\cmidrule(l){7-7}
    & NumTok          & NumBed          & +Size     & $\mathcal{L}_{R}$     & $\mathcal{L}_{C}$    & $F1$  \\ 
\midrule
    \Circled{A} & AddBack  &  CharLSTM  &  9M  &  \cmark  &  \cmark  &  \ri{86.17}{0.1} \\
    \Circled{B} & Replace  &  \grey{CharLSTM}  &  \grey{9M}  &  \grey{\cmark}  &  \grey{\cmark}  &  \rx{83.04}{0.0} \\
    \Circled{C} & \xmark  &  \xmark  &  \xmark &  \grey{\cmark}  &  \grey{\cmark}  &  \rii{85.65}{1.6} \\
    \Circled{D} & \grey{AddBack}  &  \grey{CharLSTM}  &  1M  &  \grey{\cmark}  &  \grey{\cmark}  &  \rx{84.89}{0.6} \\
    \Circled{E} & \grey{AddBack}  &  \grey{CharLSTM}  &  0.1M  &  \grey{\cmark}  &  \grey{\cmark}  &  \rx{84.01}{0.8} \\
    \Circled{X} & \xmark  &  \xmark  &  \xmark  &  \xmark  &  \grey{\cmark}  &  \rx{84.79}{0.3} \\
\bottomrule
\end{tabular}
}
\caption{LUNA Evaluations on TabBERT and CrediTrans. Since TabBERT is pre-trained from scratch, nothing can be learnt from the teacher model and $\mathcal{L}_{D}$ is not used. }
\label{tab:tabformer-eval}
\vspace{-5mm}
\end{table}

\subsubsection{Evaluation Tasks and Metrics}
First, we evaluate NumBed on toy tasks purely about numbers. We construct a pure number dataset from real-world numbers in the number pre-training dataset mentioned above. Utilizing probing models and training processes of DICE~\citep{sundararaman-etal-2020-methods}, we evaluate models' numeracy abilities on a series of number-related tasks: 1) decoding the number, 2) predicting the result of adding/subtracting two numbers, 3) locating the maximum value in a list of numbers. For 1) and 2), unlike DICE, we separately report RMSE for the significand and ACC for the exponent\footnote{\Eg, the significand and exponent of the number 3142 are 3.142 and 3 respectively, since 3142=3.142E3} instead of RMSE for the original value, because we think numbers with different scales are equally important (giving 10 when gt=1 should be worse than giving 1200 when gt=1000). 


Then, we evaluate LUNA on real-world rich-number tasks: TAT-QA, TabFact, and CrediTrans. We split the dataset into train, valid, and test sets according to their previously reported ratios: 8:1:1, 
8:1:1, and
6:2:2, respectively. In this paper, by default, the test sets will be used for reporting evaluation metrics. 

On TAT-QA, the original evaluation metric $EM$ (Exact Match) and $F1$ scores will be adopted. $EM_{num}$ is the exact match score over arithmetic questions that are officially split from TAT-QA by their authors. The arithmetic questions are those whose answer is given by operating ``$+/-/\times/\div$'' on numbers from the table or the text. On TabFact, beyond the original accuracy $ACC$, we also calculate the $ACC_{cx}$ on its complex subset. On CrediTrans, its original $F1$ score is calculated. 

\subsection{Experimental Setups}
\label{sec:exp-setup}

\subsubsection{Baseline and Ablation Studies}
To understand how each component of LUNA contributes to the overall improvements, we design a series of ablation studies. First, with part I number plugin and part II number pre-training \vs without them (\Circled{0}, \Circled{8} and \Circled{X} in \reftab{tab:luna-eval}). Here, row \Circled{X} or \Circled{3} correspond to the baseline models where LUNA is not applied or DICE is used as NumBed. Then, during pre-training, with the new losses $\mathcal{L}_{REG}$ and $\mathcal{L}_{distill}$ \vs without them (\Circled{0}, \Circled{6} and \Circled{7}). 

For fair comparisons, all evaluations are done on 1 node with the same environment configuration. By default, all evaluation metrics reported in the following are averaged over \textbf{3 runs} for experiments with randomness, and the standard error is reported along with the mean as a subscript (in the form of $mean_{std}$). Please find more training details in Appendix \refsec{app:training}.

\subsubsection{Comparing Number Plugin Choices}
\label{sec:compare-number-plugin}
There could be many possible ways to design NumTok (see \refsec{sec:numtok}) and NumBed (see \refsec{sec:numbed}). Among many, we choose several promising ones to compare through controlled experiments. For NumTok, we compare the AddBack and Replace algorithms. They correspond to row \Circled{0} and \Circled{1} respectively. For NumBed, CharLSTM \Circled{0} and CharFormer \Circled{2} are compared with each other. For part I of LUNA, the major hyper-parameter is the model size of NumBed. In column ``+Size'' of \reftab{tab:luna-eval} and \reftab{tab:tabformer-eval}, we will examine how model size impact the improvement margin on downstream tasks. Specially, NumBed choices are also discussed on toy tasks in \reftab{tab:numeracy}, where row ``RoBERTa''\footnote{encoding the number string with RoBERTa and using the pooled output as the number embedding} and ``DICE'' are baselines. 

\subsubsection{Exploring Pre-training from Scratch}
\label{sec:exp-scratch}
As we have discussed in \refsec{sec:num-pretrain}, pre-training from scratch is an option when enough computing resources and time are available. Based on our experiments of pre-training from checkpoints in \reftab{tab:luna-eval}, in \reftab{tab:tabformer-eval} we design several controlled experiments on NumTok strategies, NumBed model size and loss designs. Here, row \Circled{X} corresponds to the baseline TabBERT where LUNA is not applied. Row \Circled{C} denotes adding $\mathcal{L}_{R}$ to the baseline TabBERT but without using Number Plugins. 

Note that TabBERT is mainly designed for multivariate time series data, it models both inter- and intra-row dependencies by tokenizing rows at the cell level (\ie, each cell corresponds to a single token) and using a hierarchical structure to convert cell embeddings in the same row into row embedding.
Since each cell corresponds to a single token, the original-defined \textit{AddBack} option of NumTok is hard to apply on TabBERT while the \textit{Replace} option is still applicable. As an alternative, we introduce a new way to ``insert'' \texttt{<num ?>} in the case of TabBERT by adding the NumBed embedding back to the original vocabulary token embeddings as a variant of AddBack. 
This variant of \textit{AddBack} still tries to retain original token information while infusing additional number information. 





\subsection{Results and Insights}
\label{sec:results}

\begin{table}[t]
\centering
\resizebox{0.35\textwidth}{!}{%
\begin{tabular}{cc|cc} 
\toprule
    Subset & Count & baseline & ours \\
\midrule
    Span & \rxo{701} &  \rxo{62.62} & \rxo{63.48}  \\
    Multi-span & \rxo{217} & \rxo{68.66} & \rxo{71.43} \\
    Arithmetic & \rxo{654} & \rxo{36.39} & \rio{69.57} \\
    Count & \rxo{32} & \rxo{43.75} & \rxo{43.75} \\
\bottomrule
\end{tabular}
}
\caption{LUNA Evaluations on TAT-QA Subsets. \textmd{\small For TAT-QA, the metric is $EM$, and ``baseline'' and ``ours'' respectively denote row \Circled{X} and \Circled{0} in \reftab{tab:luna-eval}.}}
\label{tab:subset}
\vspace{-5mm}
\end{table}

By examining \reftab{tab:luna-eval} and \reftab{tab:tabformer-eval}, one can find following insights.

On all rich-number tasks, models with LUNA outperform their original ones. For each column in \reftab{tab:luna-eval} and \reftab{tab:tabformer-eval}, the top results are always achieved by applying one variation of LUNA (row \Circled{0} - \Circled{8}) and higher than the original models (row \Circled{X}) or related work (row \Circled{3}). This means by properly selecting part I and part II options for a model and a task, LUNA could help boost performance.

On TAT-QA task, the previous baseline RoBERTa is improved by more than 9\% after applying LUNA (row \Circled{0}, column ``RoBERTa'' in \reftab{tab:luna-eval}). On CrediTrans task, the SOTA model TabBERT is also improved by applying LUNA (row \Circled{A} in \reftab{tab:tabformer-eval}).

AddBack option (row \Circled{0}) for NumTok is generally a good choice comparing to the Replace option (row \Circled{1}). The gap is significant in RoBERTa and BERT results. AddBack also yields the best result (row \Circled{A}) on CrediTrans task. 

The best choice of NumBed varies for models and tasks, but CharLSTM is an overall good choice. By comparing row \Circled{0}, \Circled{2}, \Circled{3} with each other, we can find that NumBed choices matter. CharLSTM and CharFormer are significantly better than DICE, and this phenomenon occurs on the toy task~\reftab{tab:numeracy} as well. According to those experiment results, we can say that DICE is not a strong baseline for NumBed. Specially, we also run another baseline for RoBERTa on TAT-QA that replaces number strings with their language representations (\eg, replacing ``1,100'' with ``a thousand and one hundred''), and the result ($EM=33.35$, $F1=41.29$, and $EM_{num}=55.13$) is too bad to be comparable. 

For part II number pre-training of LUNA, we find that model distillation is also an overall good choice. Comparing row \Circled{0} and \Circled{7}, we find the gain of adding the distillation loss as regularization is large while no significant drawback is shown. However, the effect of regression loss on number tokens is dependent on tasks. \Eg, for TAT-QA, row \Circled{6} is worse than row \Circled{0}; for CrediTrans, row \Circled{X} is worse than row \Circled{A}; but for TabFact, row \Circled{6} is better than row \Circled{0}. This might because, TAT-QA and CrediTrans are tagging tasks that tag on each token of the last hidden state, and TabFact is a classification task that only uses the \texttt{<cls>} token of the last hidden state. \uline{When we add $\mathcal{L}_{D}$ to each masked token, the information of <cls> gets diluted and perturbed. }

The size of the NumBed model also matters. From \Circled{5} to \Circled{4} to \Circled{0}, and from \Circled{E} to \Circled{D} to \Circled{A}, with the NumBed size grows, the performances always become better. 



In detail, we also analyze the improvement we bring to different subsets of TAT-QA. We select the best random seed among all three repeats to do subset result analysis. As shown in \reftab{tab:subset}, our method mostly improves the results on arithmetic questions in TAT-QA (from 36.39 to 69.57). This might because arithmetic questions need the most numeracy capability which LUNA provides. 

Let's summarize, 
\begin{itemize}
    \item LUNA indeed helps for transformer-based language models. 
    \item For NumTok choices, AddBack is much better than Replace. 
    \item For NumBed choices, CharLSTM is an overall good choice (much better than the previously popular DICE), and enlarging numbed model size also helps. 
    \item For number pre-training, distillation loss helps when pre-training from a checkpoint, and regression loss only helps on tagging tasks. 
\end{itemize}

To figure out how LUNA works, we also do empirical studies in \refsec{sec:empirical}, including visualization of attention maps (in \refsec{sec:empirical-case}) and embeddings from different layers (in \refsec{sec:empirical-embedding}). 




%% file: 7-conclusion.tex
\label{sec:conclu}
In this paper, we propose a patch framework for language models -- LUNA (Language Understanding Number Augmentations on Transformers via Number Plugins and Pre-training) -- to enhance their numerical reasoning and calculation capabilities. Through thorough experiments and empirical studies, we show that by adding number embeddings from the whole raw number string, and continuously pre-training language models with downstream data, the performance of the language models could improve considerably. We believe that the techniques proposed in LUNA could be applied to more scenarios including Math Word Problems and other rich-number tasks.

%% file: 5-discussion.tex
LUNA is still preliminary work that requires in-depth explorations in the future. Many possible choices and evaluations are not included in the paper due to time and space limitations.

First, our current NumTok design only handles decimal numbers, ignoring edge cases such as numbers in scientific notation (\eg, ``1.5e-9''), non-decimal bases (\eg, ``0x12BF''), and formats out of arabic-hindu notations (\eg, Roman number ``XII'', natural language numbers ``twenty-one'' and ``veintiuno''). Also, units and magnitudes are not taken into the tokenizer design. Some words could even only be understood by a mixture of numbers and words (\eg, ``H2O''). All these aspects could be improved in future work.


Second, our current NumBed approach only supports encoder-only language models. In this paper, we only explore number encoders and leave out number decoders. This prevents us from applying the LUNA ideas to generative language models such as TaPeX, TableGPT, \etc Also, the best choice of NumBed and how to initialize its parameters is still an open question. How to bring the best number representing methods together still requires lots of research efforts.

Third, MLM with regression loss and model distillation is a relatively simple approach for intermediate pre-training. There are many other possible pre-training objectives to be tried for better number understanding. For example, can we design pre-training objectives for a series/column of numbers?

Finally, before proposing LUNA, we have made many failed attempts. \Eg, using the Numbed trained on toy tasks as initialization for number pre-training, or using number prompts that gives additional key and value for each transformer layer instead of directly inputting NumBed. We cannot derive rigorous proof of why they don't work. 







%% file: ethic.tex
\noindent
\textbf{Datasets}\quad
This work collects the public dataset for research perposes. We believe there is no privacy issue, because TAT-QA, TabFact, and CrediTrans are accessible to the public. 

\noindent
\textbf{Models}\quad
This work uses three large-scale language models -- RoBERTa, BERT, and TabBERT -- among which RoBERTa and BERT are PLMs pre-trained on clean and non-evil text, and all the three models are number-pretrained/finetuned on clean downstream data. Therefore, we can make sure that the models we trained will not produce discriminatory answers. 

\noindent
\textbf{Computational Resources}\quad
Our methods require low computational resources because the designed number pre-training is an intermediate pre-training that takes a few hours to be completed. Besides, the carbon footprints of all GPUs are monitored in real-time. 

%% file: appendix.tex
\section{Code and Data}

Please find our current code and data at \url{https://github.com/zmy/LUNA}


\section{Number Statistics}
\label{sec:number-stat}

In TAT-QA and TabFact datasets, 100\% and 94.79\% of the tables have numbers in them. In our number pre-train dataset, 97\% of the tables have numbers in them, 54.94\% of all the 2,471,520 table cells have numbers in them. (While only 17.72\% table cells contain at least 2 numbers.) When analyzing tables at row-level or column levels, numbers are also hard to ignore for 71.36\% of the table columns and 85.25\% of the table rows.

Numbers are shown as strings in tables and texts. After ignoring edge cases, such as scientific notation, the remaining decimal number strings have clear format patterns in them: Most of them are positive numbers (0.12\% of all numbers are negative); 91.00\% and 9.00\% of all numbers are integers and float numbers, respectively; 1.51\% and 5.02\% of all numbers have percent ``\%'' or comma ``,'' character in them, respectively. 

Compared to tables, in plain texts numbers are relatively sparse. Out of 59,645 pieces of texts (including queries, paragraphs, and captions) in the three datasets, there are 934,121 English words but there are only 76,532 numbers in them.

\section{NumTok Details}
\label{app:numtok}


\subsection{NumTok Character Set}
\label{app:numtok_vocab}
In this work, we chose a set of characters that suits our downstream tasks the best, which consists of numeric digits 0-9, the percentage symbol (``\%''), the plus and minus signs (``+'', ``-''), the decimal point (``.''), and the thousands separator (``,''). This set can also be updated to suit different downstream tasks.

\subsection{Number Regular Expressions}
\label{app:numtok_re}
In this work, we designed regular expressions for integral, float, percentage and thousands-separated numbers.
\begin{table}[h]
\centering
\resizebox{0.45\textwidth}{!}{%
\begin{tabular}{cc}
    \toprule
    Number shape & Regular~Expression\\
    \hline
    Conventional & [+-]?$\backslash$d+(?:$\backslash$.$\backslash$d*)?\%? \\
    Thousands-separated & [+-]?$\backslash$d\{1,3\}(?:,$\backslash$d\{3\})*(?:$\backslash$.$\backslash$d+)?\%? \\
    Dot-started Decimal & [+-]?$\backslash$.$\backslash$d+\%? \\
    \bottomrule
\end{tabular}
}
\caption{NumTok Regular Expressions}
\end{table}


\section{Training Details}
\label{app:training}
\begin{table}[t]
    \centering
    \includegraphics[width=\linewidth]{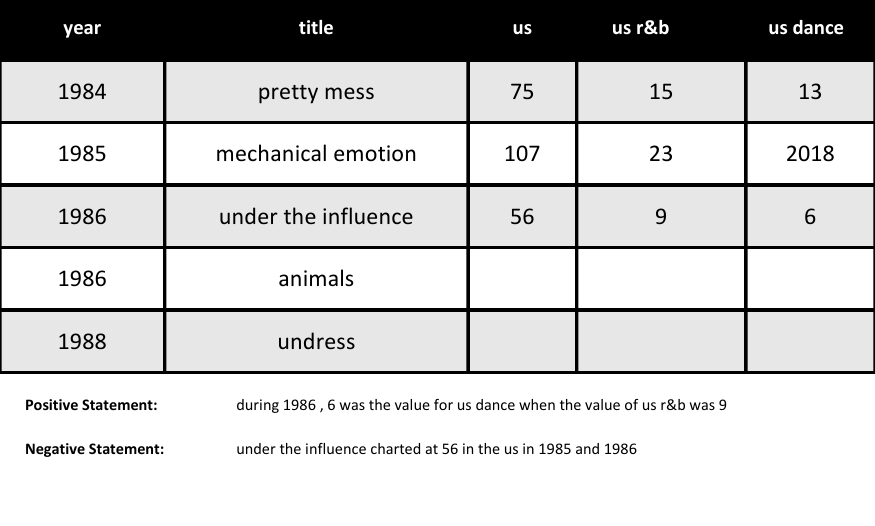}
    \caption{TabFact Example: A Wikipedia Table.}
    \label{tab:B}
\end{table}

\begin{table}[t]
    \centering
    \includegraphics[width=\linewidth]{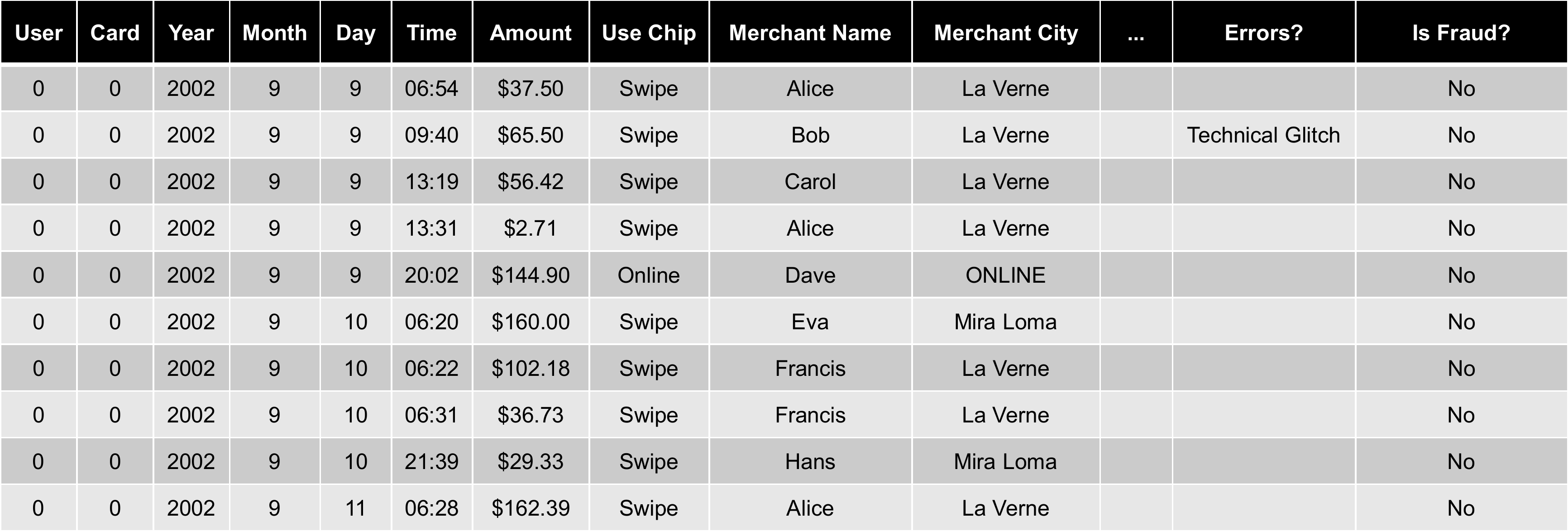}
    \caption{CrediTrans Example: A Transaction Table.}
    \label{tab:C}
\end{table}

\subsection{Number Pre-training Datasets}
\label{app:number-datasets}
 We first construct a number pre-training dataset (TAT-QA+WikiTables) for RoBERTa and BERT (pre-training from checkpoints). For TAT-QA, questions and paragraphs will be equally treated as text, and for TabFact, only positive statements of a table from the training set (to avoid data leakage) will be considered as text. Totally, 103K table-text pairs are collected. The collected dataset contains 18K tables. Over these tables, the average count of rows and columns is 14.5 and 6.8 respectively, and the average number of non-empty cells is 30.2. The average number of words in the text is 19.3. (see \refsec{sec:number-stat} for more statistics. ) 
 
 For TabBERT (pre-training from scratch), following \citep{padhi2021tabular}, we create samples by combining 10 contiguous rows in a time-dependent manner and quantize continuous fields to build a local finite vocabulary. All 2.4M samples are used in pre-training. To avoid data leaking for the downstream fraud detection task, we exclude the label column "isFraud?" in the pre-training corpus.
 
\subsection{RoBERTa}
For number pre-training, we set lr=4e-5, epoch=4, batch-size=96, and totally train 4.2K steps on 32 GPUs. 
For TAT-QA finetune, we set lr=5e-6 for RoBERTa and lr=1.5e-4 for the rest of parameters, epoch=25, batch-size=32, and totally train 10K steps on 8 GPUs. 

\subsection{BERT}
For number pre-training, we set lr=3e-5, epoch=18, batch-size=24, and totally train 77K steps on 8 GPUs. 
For TabFact finetune, we set lr=1e-5, epoch=10, batch-size=48, and totally train 19K steps on 8 GPUs. 

\subsection{TabBERT}
For pre-training, following the original implementation of TabBERT, we set lr=5e-5, epoch=3, batch-size=24, and totally train 38K steps on 8 GPUs. For CrediTrans finetune, we set lr=1e-3, epoch=10, batch-size=256. During the fine-tuning stage, The TabBERT model is fixed as a feature extractor and we only update the parameters of the LSTM prediction head.






\section{Empirical Studies}
\label{sec:empirical}
In this section, we select several cases in downstream tasks to demonstrate how LUNA improves number understanding. Then we visualize the number embedding space learned by NumBed, and compare it with the subword token embedding space.

\subsection{Case Studies}
\label{sec:empirical-case}
LUNA improves language models by helping them fixing number-related errors. In \reftab{tab:subset} we have already shown the statistics. Now let's take a look at how the errors discussed in \refsec{sec:tab-models} are fixed with the help LUNA framework.
\begin{figure}[t]
    \centering
    \includegraphics[width=0.5\textwidth]{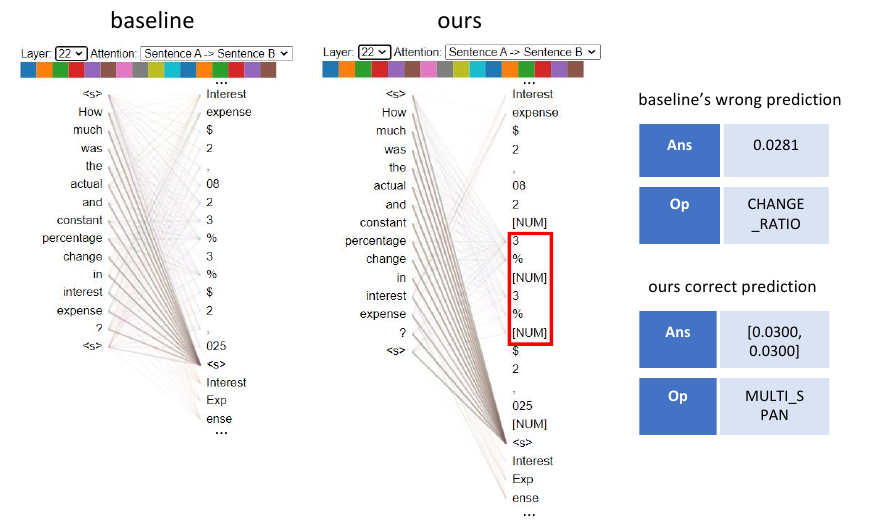}
    \caption{A part of attention map between question and table+paragraph of \reftab{tab:A}. ``baseline'' and ``ours'' respectively denote row \Circled{X} and \Circled{0} in \reftab{tab:luna-eval}. }
    \label{fig:tat-attention}
\end{figure}

For the TAT-QA input case mentioned in \reftab{tab:A}, the original Roberta tags at ``\$2,082'' and ``\$2,025'', operates ``CHANGE\_RATIO'' to them and returns ``0.0281''. This is because, the model is confused by the segment, ``percentage change in interest expense'', in the question. The error shows that the model cannot understand numbers and correlate numbers to text. When seeing the word ``percentage change'', the model arbitrarily filters the same-number pairs. 

\reffig{fig:tat-attention} shows the attention map of the last but one transformer layer of RoBERTa with \reftab{tab:A} as input. As you can see, the original attention map is a mess, which leads to the wrong prediction. Meanwhile, our method focuses on the core information, ``3\%'', which results in the correct answer. 



\subsection{Number Embedding}
\label{sec:empirical-embedding}

\begin{figure}[t]
    \centering
    \includegraphics[width=0.45\textwidth]{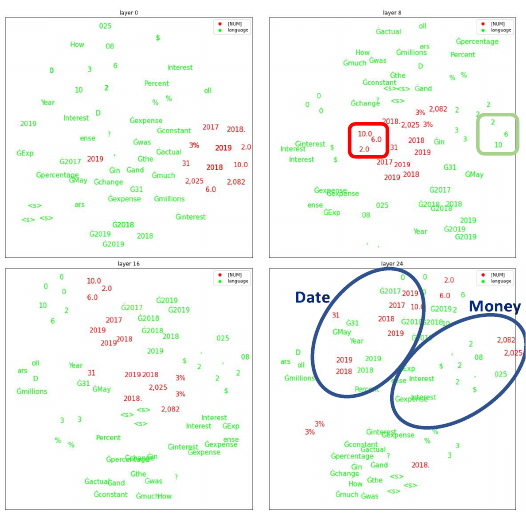}
    \caption{The TSNE visualization of the output of different layers when \Circled{0} handles \reftab{tab:A}. Note that the same number strings may have different embeddings due to their different positions in the serialized sequence. }
    \label{fig:tat-tsne}
\end{figure}

\begin{figure}[t]
    \centering
    \includegraphics[width=0.45\textwidth]{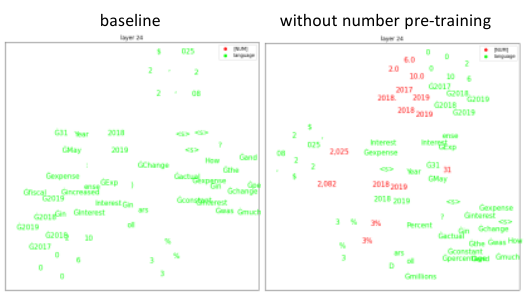}
    \caption{The TSNE visualization of other rows in \reftab{tab:luna-eval}}
    \label{fig:tat-tsne-others}
\end{figure}

We apply TSNE visualization to different transformer layers when taking \reftab{tab:A} as input. As you can see in \reffig{fig:tat-tsne}, the deeper the layer is, the better fusion between numbed and word embedding occurs. Especially in the last layer, these embeddings cluster according to their concept. For example, ``2019'', ``2018'', ``Year'', ``May'' and ``31'' belong to the concept of date. ``2,082'', ``2,025'', ``08'', ``025'', ``\$'' and ``Expense'' mean the money. On the other hand, in shallow layers like layer 8, low-level and fine-granted knowledge is learned. For instance, the NumBed ``2.0'', ``6.0'', and ``10.0'' are close to each other but the word embedding ``2'', ``6'', and ``10'' are kind of far away. 

\reffig{fig:tat-tsne-others} shows the TSNE visualization of another two rows in \reftab{tab:luna-eval}. As shown in the left sub-figure, the cluster is not obvious without LUNA. In the right sub-figure, without intermediate pre-training, the distance between ``2,082'' and ``2,025'' is longer than row \Circled{0}, which means the model cannot understand the numbers as well as row \Circled{0}.  



